\documentclass{article}
\usepackage{ijcai21}

\usepackage{times}

\usepackage{soul}
\usepackage{url}
\usepackage[hidelinks]{hyperref}
\usepackage[utf8]{inputenc}
\usepackage[small]{caption}
\usepackage{graphicx}
\usepackage{amsmath}
\usepackage{booktabs}
\title{Mapping Slums with Medium Resolution Satellite Imagery: a Comparative Analysis of Multi-Spectral Data and Grey-level Co-occurrence Matrix Techniques}

\author{
Agatha C. H. de Mattos\footnote{Contact Author}\and
Gavin McArdle\and
Michela Bertolotto\\
\affiliations
School of Computer Science, University College Dublin\\
\emails
agatha.hennigendemattos@ucdconnect.ie,
\{gavin.mcardle, michela.bertolotto\}@ucd.ie
}

\begin{document}

\maketitle

\begin{abstract}
The UN-Habitat estimates that over one billion people live in slums around the world. However, state-of-the-art techniques to detect the location of slum areas employ high-resolution satellite imagery, which is costly to obtain and process. As a result, researchers have started to look at utilising free and open-access medium resolution satellite imagery. Yet, there is no clear consensus on which data preparation and machine learning approaches are the most appropriate to use with such imagery data. In this paper, we evaluate two techniques (multi-spectral data and grey-level co-occurrence matrix feature extraction) on an open-access dataset consisting of labelled Sentinel-2 images with a spatial resolution of 10 meters. Both techniques were paired with a canonical correlation forests classifier. The results show that the grey-level co-occurrence matrix performed better than multi-spectral data for all four cities. It had an average accuracy for the slum class of 97\% and a mean intersection over union of 94\%, while multi-spectral data had 75\% and 64\% for the respective metrics\footnote{Code available at: \href{https://github.com/ml-labs-crt/slums-med-res}{https://github.com/ml-labs-crt/slums-med-res}}. These results indicate that open-access satellite imagery with a resolution of at least 10 meters may be suitable for keeping track of development goals such as the detection of slums in cities.
\end{abstract}
\section{Introduction}
\noindent It is estimated that over one billion people around the world live in slums \cite{habitatTrackingProgressInclusive2018}. A slum, according to United Nations Habitat, is a household where residents lack at least one of the following: water, sanitation, housing durability, security of tenure or sufficient living area \cite{habitatTrackingProgressInclusive2018}. However, other definitions of what constitute a slum exist \cite{lilfordBecauseSpaceMatters2019} and some scholars argue that a better definition would remove lack of security of tenure or sufficient living area, as these conditions are not exclusive to slums \cite{doveyViewpointInformalSettlement2020}. Still, despite the criticism about what would be the most appropriate definition for slums, Goal 11.1 of the Sustainable Development Goals states that adequate, safe and affordable housing should be accessible for all and that slums should be upgraded by 2030 \cite{unitednationsTransformingOurWorld2015}. Countries are therefore expected to be able to map slums, quantify the population living in these settlements and to ensure that high-quality, accessible, timely and reliable disaggregated data is made available at national and sub-national levels \cite{habitatTrackingProgressInclusive2018}. However, keeping up-to-date records of the slum areas can be challenging. Currently, the main source of information about the percentage of the world urban population living in the slums comes from housing census surveys, which are labour-intensive, time consuming, and require substantial financial resources \cite{mahabirCriticalReviewHigh2018}. As a result, census data are only reported at lengthy intervals that can go from five, ten or more years, causing the information to become obsolete and not suitable to capture the dynamism of slums \cite{mahabirCriticalReviewHigh2018}. An additional challenge comes with the reluctance of some governments to recognize slums, so they are often excluded from official statistics \cite{habitatTrackingProgressInclusive2018}.\par
Consequently, there is an interest from many stakeholders to reduce or combine the use of field surveys with other forms of data collection \cite{mahabirCriticalReviewHigh2018,gram-hansenMappingInformalSettlements2019}. Given the physical nature of slums and large volume of data required to keep track of the development of slums at a global scale, researchers have investigated approaches employing satellite imagery and image processing techniques for the task of mapping these settlements \cite{kufferSlumsSpace152016,mahabirCriticalReviewHigh2018}. Such automated techniques have the advantage of reducing costs, increasing the frequency of updates and potentially tackling under-reporting and quality issues that may arise in the census collection process \cite{mahabirCriticalReviewHigh2018,leonitaMachineLearningBasedSlum2018} .\par
While progress has been made over the years, slum detection is still an open problem that has received very little attention from the AI community. A review of studies published between 2000 and 2015 shows that only 12.6\% employed machine learning \cite{kufferSlumsSpace152016}. Though more research has been conducted since then, the majority of the studies focus on high or very high-resolution imagery \cite{wurmExploitationTexturalMorphological2017,gram-hansenMappingInformalSettlements2019}. High-resolution images provide richer information about the areas, but it is also associated with a higher acquisition and computation costs. It also makes large country-wide or worldwide analysis less feasible, something that has been argued to be desirable and to be a limitation of current research \cite{kufferSlumsSpace152016}.\par
Another issue is the lack of ground-truth data. Because official census data are often made available at aggregated national levels, the data can not be directly used to train models. Most researchers opt for generating their own ground truth labels \cite{vermaTransferLearningApproach2019}, engaging in field surveys \cite{kufferExtractionSlumAreas2016} or partnering with institutions \cite{gram-hansenMappingInformalSettlements2019,leonitaMachineLearningBasedSlum2018,kufferExtractionSlumAreas2016}. As a result, very few researchers make their ground truth data available which invariably hinders the development of research in this area. To the best of our knowledge, only Gram-Hansen et al. \shortcite{gram-hansenMappingInformalSettlements2019}, Bayle and Silvani \shortcite{bayleCaseStudyMapping2020} and Verma et al. \shortcite{vermaTransferLearningApproach2019} have made their ground-truth labelled data available.\par
The high cost of acquiring high-resolution imagery and lack of ground-truth data has led to a situation in which techniques developed for detecting slums were tested on different imagery, captured from different satellites and with a different resolution, for distinct locations and often reported using diverse metrics. In this paper, we sought to fill this gap in the literature by evaluating the performance of two techniques on the same locations using free-accessible medium resolution satellite imagery and ground-truth data. We compared the techniques with regards to the pre-processing of the satellite images, computing time, number of parameters that need to be set, and performance.\par
The remainder paper is organized as follows: section 2 describes related work, section 3 introduces the methodology used in this study; and section 4 presents the results and a discussion. Section 5 concludes the paper and suggests possible future work. \par
\section{Related Work}
\noindent Most of the research in slum mapping to date uses remote sensing, specially high-resolution and very high-resolution satellite imagery. In this context, very high-resolution refers to images with a spatial resolution of less than 4 meters, often below one meter, high-resolution to resolutions between 4 and 10 meters and medium resolution for values between 10 and 50 meters. For images with a high resolution, a great diversity of techniques have been tested and the reviews made by Kuffer et al.  \shortcite{kufferSlumsSpace152016} and Mahabir et al. \shortcite{mahabirCriticalReviewHigh2018} present an overview of the research field until 2016. Since then, some researchers have investigated how these different methods compare when applied to the same dataset. For example, studies have looked into different features that can be extracted from images \cite{duqueExploringPotentialMachine2017}, the most appropriate choice of parameters when extracting features \cite{kufferExtractionSlumAreas2016,leonitaMachineLearningBasedSlum2018} and the impact of the classifier that is paired with these features \cite{duqueExploringPotentialMachine2017,leonitaMachineLearningBasedSlum2018,gadirajuMachineLearningApproaches2018}. More recently, scientists have examined how approaches that use patches (or tiles) of an image, such as convolutional neural networks, compare with feature extraction and single pixel classifiers \cite{mbogaDetectionInformalSettlements2017,gadirajuMachineLearningApproaches2018}. \par
Research on slum mapping using medium-resolution imagery is scarce and restricted to a handful of articles. Wurm et al. \shortcite{wurmExploitationTexturalMorphological2017} investigated how the parameters set for grey-level co-occurrence matrix and differential morphological profiles feature extraction impacted the detection of slums when paired with a random forest classifier and Gram-Hansen et al. \shortcite{gram-hansenMappingInformalSettlements2019} used multi-spectral data paired with canonical correlation forests to map slums. Though both researchers used medium resolution imagery coming from the same satellite (Sentinel-2), their results are not comparable because they were applied to different cities using a distinct methodology. Wurm et al. \shortcite{wurmExploitationTexturalMorphological2017} investigated their approach in Mumbai, India while Gram-Hansen et al. \shortcite{gram-hansenMappingInformalSettlements2019} tested theirs in ten locations around the globe that have slums. Additionally, Wurm et al. \shortcite{wurmExploitationTexturalMorphological2017} pre-processed the images for atmospheric distortions, a step that was skipped by Gram-Hansen et al. \shortcite{gram-hansenMappingInformalSettlements2019}. In a similar manner, Gram-Hansen et al. \shortcite{gram-hansenMappingInformalSettlements2019} trained their models on a balanced dataset, a step that was not taken by Wurm et al. \shortcite{wurmExploitationTexturalMorphological2017}. Gram-Hansen et al. \shortcite{gram-hansenMappingInformalSettlements2019} and Wurm et al \shortcite{wurmExploitationTexturalMorphological2017} also used a distinct split of the train/test set - 80/20 and 20/80, respectively - something that can impact a model’s ability to learn to distinguish the intra-variability of the slum class \cite{starkSlumMappingImbalanced2019}. Finally, both works employed a different classifier, random forest and canonical correlation forests, with different parameters which also yields different results.\par
Our work aims to provide an objective comparison of these two state-of-the-art techniques. By controlling for the test and ground truth data (the locations) and the classifier, a fair comparison of the techniques is possible. In the next section, the methodology used in our work is described.\par
\section{Methodology}
\begin{figure}
\centering
\includegraphics[width=0.85\columnwidth]{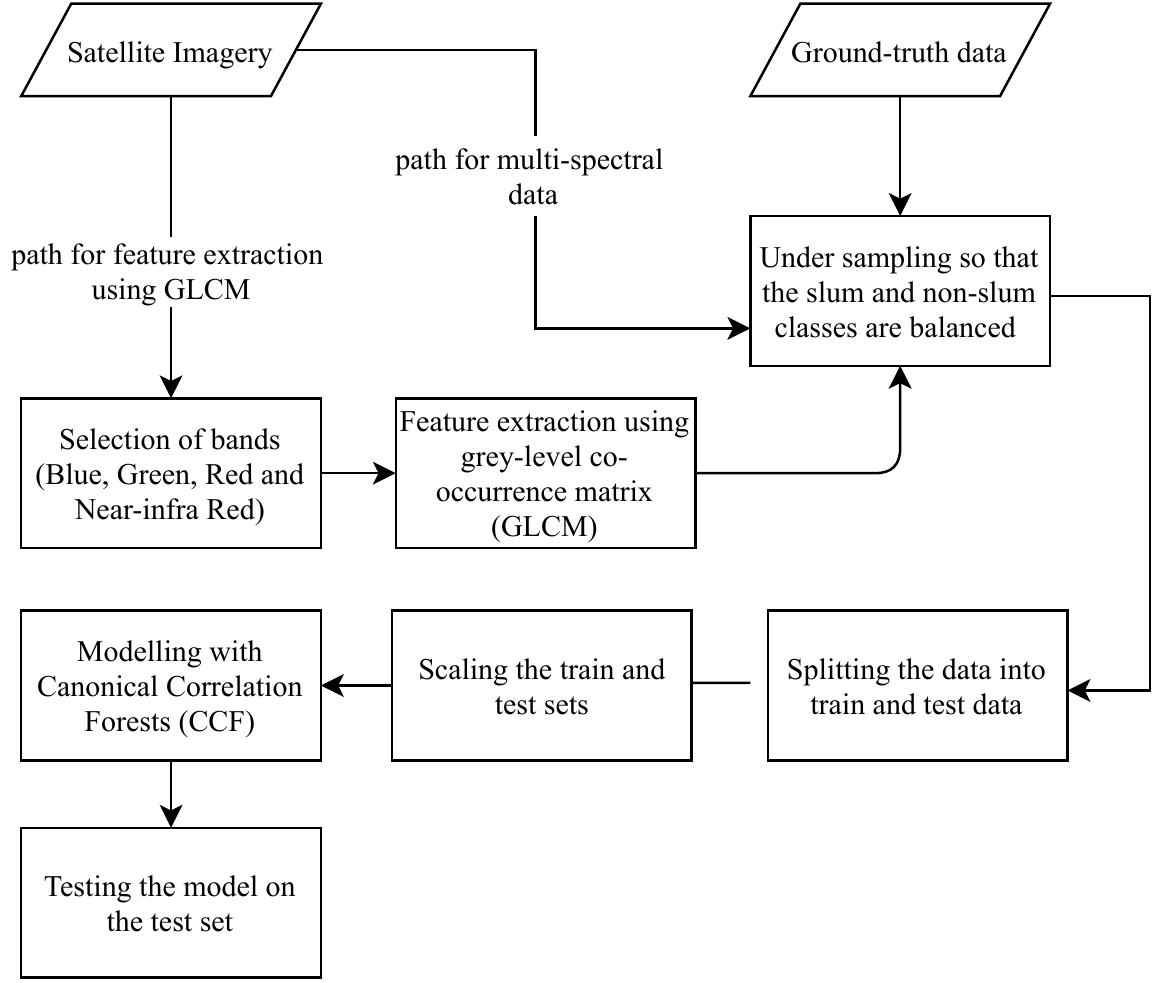} 
\caption{Methodology used to compare Techniques using Multi-Spectral Data and Grey-Level Co-occurrence Matrix.}
\label{diagram}
\end{figure}

\noindent The aim of this research is to fairly compare the techniques presented by Gram-Hansen et al. \shortcite{gram-hansenMappingInformalSettlements2019} and Wurm et al. \shortcite{wurmExploitationTexturalMorphological2017} for slum detection using medium resolution images. To allow a fair comparison some changes to the original methodologies were required. For instance, images were not corrected for interactions with the atmosphere, class imbalance was addressed and features were paired with Canonical Correlation Forests (CCFs) for classification. Figure \ref{diagram} summarizes the methodology. Details of the datasets used, pre-processing of the data, methods employed, parameters selected and evaluation criteria are presented below.\par
\subsection{Datasets}
\noindent The dataset chosen for this experiment was made available by Gram-Hansen et al. \shortcite{gram-hansenMappingInformalSettlements2019} and is composed of satellite images and ground-truth data of slum locations in six countries. The images are available in both high resolution and medium resolution, but only the latter was used, as it is the intent of this paper to compare techniques that can be applied for this type of imagery. The images were obtained by the European Space Agency through the Sentinel-2 mission and are recorded using 13 bands. Each band of data provides a record of the amount of energy reflected in a specific portion of the electromagnetic spectrum recorded by the satellite's sensors. This information is stored and converted to picture format. Together the bands constitute what is called multi-spectral data. Not all bands are recorded with the same spatial resolution and bands with lower resolution tend to be more affected by the atmosphere \cite{gram-hansenMappingInformalSettlements2019}. With that in mind, Gram-Hansen et al.\shortcite{gram-hansenMappingInformalSettlements2019} removed bands 1, 9 and 10 from the data as these bands are provided at lower resolutions and, as a consequence, interact more heavily with the atmosphere. The remaining 10 bands - 2, 3, 4, 5, 6, 7, 8, 8A, 11, 12 - were kept as they are provided at either 10m or 20m spatial resolution and hence are less affected by the atmospheric distortions \cite{gram-hansenMappingInformalSettlements2019}. \par
The ground-truth is composed of images with the same size as the images described above, but that have been annotated with one and zero labels representing the binary classification of slum and non-slum. According to the creators of the dataset, the ground-truth was partially obtained in partnership with other organizations and partially annotated by the researchers themselves \cite{gram-hansenMappingInformalSettlements2019}.\par
Gram-Hansen et al. \shortcite{gram-hansenMappingInformalSettlements2019} tried to ensure that the images and annotations were aligned in space, but there are still some small differences. Of all locations, four were selected for this experiment as they did not have any null values in the ground-truth data and were ready for analysis. The locations were El Daein and Al Geneina in Sudan, Makoko in Nigeria and Medellín in Colombia. 
\subsection{Pre-Processing, Methods, Parameters Selection, and Evaluation}
\noindent The datasets are imbalanced and the percentage of pixels classified as slums can vary from 22\% to 43\%. This has been shown to interfere with a model’s ability to learn to distinguish the intra-variability of the slum class and to change results \cite{starkSlumMappingImbalanced2019}, so for both techniques the train and test sets were balanced by undersampling the majority class (see Figure \ref{diagram} for the pre-processing steps). In the case of Medellín, that class was the slum class, but for all other three locations, it was the non-slum class. Through the experiment, whenever it was required to random sample, the seed was set to zero. The training set contained 80\% of the data points available and the test set 20\%. The training set was scaled before classification in both experiments so that the mean of data used to train the model was zero and the standard deviation one. The test set was scaled using the training set mean and standard deviation. All analyses were implemented using R version 4.0.2 on Windows 10 using an Intel Core i7-8665U CPU equipped with memory of 16 GB. The code for this experiment is available at: \href{https://github.com/ml-labs-crt/slums-med-res}{https://github.com/ml-labs-crt/slums-med-res}.\par
\textbf{Canonical Correlation Forests (CCFs)} is a tree ensemble method that employs canonical correlation analysis and projection bootstrapping during the training of each tree \cite{rainforthCanonicalCorrelationForests2017}. Similarly to Gram-Hansen et al. \shortcite{gram-hansenMappingInformalSettlements2019}, the number of trees used in our experiments was 10. In all analyses, the CCF R package 0.1.0 was used \cite{doblerCanonicalCorrelationForest2019}.\par
\textbf{Grey-Level Co-occurrence Matrix (GLCM)} was described by Haralick \shortcite{haralickTexturalFeaturesImage1973} and it is a well-known technique used in image processing to extract texture features from an image. Most parameters were set following the work of Wurm et al. \shortcite{wurmExploitationTexturalMorphological2017}, with exception of the number of grey levels, that was not discussed in their paper. This parameter was set to 32, as grey-levels greater than 24 are advocated and greater than 64 are deemed unnecessary as they do not improve classification accuracy \cite{clausiAnalysisCooccurrenceTexture2002}. The other parameters were set to: window size of 19x19; spatial relationships between pixels of 0°, 45°, 90° and 135°; features were extracted for bands 2 (Blue), 3 (Green), 4 (Red) and 8 (near Infra-Red); seven feature measures were calculated, namely second moment, contrast, correlation, homogeneity, entropy and mean as well as variance. In all analyses the R package GLCM 1.6.5 was used \cite{zvoleffCalculateTexturesGreylevel2021}.\par
The output of the two approaches described above were compared using pixel accuracy, Intersection over Union (IoU), and mean IoU; three metrics that are commonly used in semantic segmentation analysis  \cite{longFullyConvolutionalNetworks2015}. 
\section{Results and Discussion}
\noindent The comparative analysis of multi-spectral data and grey-level co-occurrence matrix texture features (GLCM) paired with canonical correlation forests shows that the latter performed better for all datasets (Table \ref{tableResults}). While the average mean IoU was between 58.0\% and 66.1\% for multi-spectral data, it was between 92.7\% and 95.9\% for features extracted using GLCM. This suggests that extracting textural features provides a better representation of the data for this task than utilising the pure values of pixels. This result is in line with the results found by \cite{kufferExtractionSlumAreas2016} for very high resolution imagery.  
\begin{table*}[ht]
\begin{tabular}{c|cc|ccc|cc|ccc}
%\cline{2-11}
                                & \multicolumn{5}{c}{Multi-Spectral Data + CCF}                                     & \multicolumn{5}{|c}{Grey-Level Co-occurrence Matrix (GLCM) + CCF}                         \\ %\cline{2-11} 
                                & \multicolumn{2}{c|}{Accuracy} & \multicolumn{3}{c|}{Intersection over Union (IoU)} & \multicolumn{2}{c|}{Accuracy} & \multicolumn{3}{c}{Intersection over Union (IoU)} \\ 
\multicolumn{1}{c|}{Location}  & Slum        & Non-Slum        & Slum         & Non-Slum         & Mean IoU         & Slum        & Non-Slum        & Slum          & Non-Slum         & Mean IoU        \\ \hline
\multicolumn{1}{c|}{Medellín}  & 79.6        & 79.5            & 66.1         & 66.1             & 66.1             & 95.2        & 97.9            & 93.3          & 93.4             & 93.3            \\ 
\multicolumn{1}{c|}{Makoko}    & 66.1        & 80.9            & 55.5         & 60.4             & 58.0             & 96.8        & 95.7            & 92.8          & 92.7             & 92.7            \\ 
\multicolumn{1}{c|}{Al Geneina} & 75.7        & 82.2            & 64.3         & 66.2             & 65.2             & 97.8        & 98.0            & 95.9          & 95.9             & 95.9            \\ 
\multicolumn{1}{c|}{El Daein}   & 78.5        & 79.8            & 65.3         & 65.7             & 65.5             & 97.6        & 97.1            & 94.9          & 94.9             & 94.9            \\ 
\end{tabular}
\caption{Results of the Comparison of Multi-Spectral Data and GLCM paired with Canonical Correlation Forests (CCF).}
\label{tableResults}
\end{table*}
Although it might be expected that GLCM + CCF would take longer than multi-spectral data + CCF, the differences between them were small. For the largest dataset, the city of El Daein in Sudan, multi-spectral data + CCF took 14.3 minutes and GLCM + CCF took 15.7 minutes, only 10\% more. When considering the total time required to run the four datasets, GLCM + CCF only took 18\% longer than multi-spectral data + CCF. In fact, the biggest differences in computing time were related to the different sizes of the datasets. When it comes to the computing time of the GLCM, it is greatly influenced by the number of grey levels chosen for the analysis \cite{wurmSlumMappingPolarimetric2017}. While the image had initially 2ˆ16 levels (Sentinel-2 images are recorded using 16 bits), the reduction to 32 levels significantly reduces the size of the grey-level co-occurrence matrices which has a direct influence on computation time. Another factor to consider is the number of bands for which features were extracted. As mentioned earlier, in this paper we extracted features for four bands following from the work of Wurm et al. \shortcite{wurmExploitationTexturalMorphological2017}.\par
Though the accuracy of our experiments was high, that may be related to the train/test set proportion. As mentioned by Leonita et al. \shortcite{leonitaMachineLearningBasedSlum2018}, in slum mapping problems, it makes more sense to assign a higher proportion of the data as test data and train the model with 20\% or 30\% of the data only. Additionally, it is desirable that the training data comes from only a specific area of the city, instead of being randomly sampled from the whole image, as this would be closer to the labelled data available for most cities around the world. Ideally the experiments would have repeated with different random splits of the data as well. The results achieved using multi-spectral data are very close to those found by Gram-Hansen et al. \shortcite{gram-hansenMappingInformalSettlements2019}. Any differences might be attributable to the different platforms and packages used (MATLAB vs. R) and the sampling technique used to balance the datasets.\par
Interestingly, the results suggest that mean over union above 92.7\% can be found even when satellite imagery is not pre-processed for atmospheric correction which may indicate that this is a step that could be overlooked. This is in line with the work of Gram-Hansen et al. \shortcite{gram-hansenMappingInformalSettlements2019} and in contrast to the work presented by Wurm et al. \shortcite{wurmExploitationTexturalMorphological2017}, which explicitly applied techniques to correct for atmospheric interactions. \par
Perhaps the main disadvantage of GLCM is that it requires several parameters to be set: number of grey levels, direction of pixel computation, size of the window and feature measures. These parameters can either be set by trial and error or by using guidelines, such as those described by Clausi \shortcite{clausiAnalysisCooccurrenceTexture2002}. This is a disadvantage of the method, because the user does not know at the time of extraction of the parameters how they are going to perform later in the classification \cite{mbogaDetectionInformalSettlements2017}. Moreover, the variable values obtained in the GLCM approach in our test cities may not be applicable to other cities/contexts, so even though the models had a mean IoU of over 92.7\% for all analyzed cities, they may still need to be re-trained before development in other locations. \par
Regarding possible biases in the ground-truth dataset, some of the criteria used to define slums cannot be seen on satellite imagery, such as tenure of the land, so there is an intrinsic limitation in using images to detect these places depending on the definition and understanding that is given to the term “slum”. In fact, it is not known what criteria was adopted by the creators of the dataset. Moreover, the approaches discussed in this paper are (quite literally) “top-down”. In that respect, an alternative would be to engage residents through the use of volunteer georeferenced information \cite{mahabirCriticalReviewHigh2018}. 
\section{Conclusions and Future Work}
\noindent In this work we advanced the research that aims at providing up-to-date maps of the urban population living in slums by comparing the performance of multi-spectral data and grey-level co-occurrence matrix (GLCM) feature extraction techniques using medium resolution satellite imagery. Medium resolution satellite imagery allows for a less detailed observation of the city when compared to high-resolution and very-high resolution imagery, but it is freely accessible and frequently updated, two important characteristics to keep slum maps up to date. By using the same test data and classifier it was possible to meaningfully compare the results of using these two techniques. Such controlled comparisons are absent from the literature.\par
The results show that both techniques, though especially GLCM due to its higher accuracy and intersection over union score, managed to detect slums. Moreover, both methods were successfully run on commodity hardware, an important factor to make a global slum inventory possible. In this regard, though GLCM performed better than multi-spectral data, this performance came at the cost of higher computational time. Although the code to implement the techniques described in the paper, software to run the code and the datasets used are freely available online, we recognize that technical knowledge is required to use our work, which can hinder its usability outside the AI community. \par
In the future, we would like to  include locations in Southeast Asia, where most of the population living in slums in the world is located \cite{habitatTrackingProgressInclusive2018}, test the performance of the techniques paired with other classifiers and expand the comparison to include deep learning approaches such as the work of Wurm et al. \shortcite{wurmSemanticSegmentationSlums2019} and Verma et al. \shortcite{vermaTransferLearningApproach2019}.\par
Within this research, it is important to be cognisant of the choice of the word “slum”. While it can be considered a derogatory term, we used it here with the hope that our introduction helped unpacking the particular conditions to which it refers and not without failing to acknowledge its shortcomings. A more complete discussion on the pitfalls of this terminology is provided by Mayne \shortcite{mayne2017slums} and Dovey et al. \shortcite{doveyViewpointInformalSettlement2020}. Lastly, for research on slum mapping to promote positive social impact, those who have access to their output must have the interest of the poor in mind when making decisions and designing policies that will impact them.  
 
\section*{Acknowledgments}
This publication has emanated from research supported in part by a grant from Science Foundation Ireland under Grant number 18/CRT/6183. For the purpose of Open Access, the author has applied a CC BY public copyright licence to any Author Accepted Manuscript version arising from this submission.

\bibliographystyle{named}
\bibliography{ijcai2021}
\end{document}